\documentclass[journal,12pt,onecolumn]
{IEEEtran}

\ifCLASSINFOpdf
\else
\fi

\usepackage{amsmath}
\usepackage{setspace}
\usepackage{graphicx}
\usepackage{xcolor,colortbl}
\usepackage{multirow}
\usepackage{multirow,tabularx}
\usepackage[space]{cite}
\usepackage{algorithmic}
\usepackage{amssymb}
\usepackage{pifont}

\usepackage{array}
\usepackage{etoolbox}

\patchcmd{\IEEEbiographynophoto}
  {\vskip 4\baselineskip plus 1fil minus 0\baselineskip}
  {\par\addvspace{0.75\baselineskip}}
  {}
  {\PackageError{manuscript}{Could not compact biography spacing}{}}

\begin{document}

\title{Learning-Based System Identification for DVL-Constrained AUV Localization}

\author{Mingxi Zhou,~\IEEEmembership{Member,~IEEE},
Emir Cem Gezer,
Lin Zhao,~\IEEEmembership{Member,~IEEE},
William McConnell,
and Chengzhi Yuan,~\IEEEmembership{Member,~IEEE}
\thanks{Received 20 June 2023; revised 18 June 2025, 6 November 2025, 22 April 2026, and 14 August 2026; accepted 14 August 2026. This work was supported in part by the National Science Foundation under Grant 2221676 and Grant 2238168 and in part by the University of Rhode Island. (Corresponding author: Mingxi Zhou.)}
\thanks{Associate Editor: N. A. Cruz.}
\thanks{Mingxi Zhou is with the Graduate School of Oceanography, University of Rhode Island, Kingston, RI 02881 USA (e-mail: mzhou@uri.edu).}
\thanks{Emir Cem Gezer is with the Department of Marine Technology, Norwegian University of Science and Technology, NO-7491 Trondheim, Norway (e-mail: emir.cem.gezer@ntnu.no).}
\thanks{Lin Zhao was with the Department of Ocean Engineering, College of Engineering, University of Rhode Island, Kingston, RI 02881 USA (e-mail: linzhao@uri.edu).}
\thanks{William McConnell was with the Department of Mechanical Industrial and Systems Engineering, College of Engineering, University of Rhode Island, Kingston, RI 02881 USA (e-mail: wmcconnell@uri.edu).}
\thanks{Chengzhi Yuan is with the Department of Mechanical Industrial and Systems Engineering, College of Engineering, University of Rhode Island, Kingston, RI 02881 USA (e-mail: cyuan@uri.edu).}
\thanks{Digital Object Identifier 10.1109/JOE.2026.3733634}}

\markboth{IEEE Journal of Oceanic Engineering}{Zhou \MakeLowercase{\textit{et al.}}: Learning-Based System Identification for DVL-Constrained AUV Localization}

\maketitle

\begin{abstract}
Nowadays, most AUVs utilize Doppler velocity logs (DVLs) and inertial measurement units (IMUs) to achieve satisfactory dead-reckoning results. However, the DVL bottom lock may not be available in some scenarios, such as moving over a soft bottom or traveling further away from the seafloor beyond the DVL's range. As a result, the localization system solely reliant on the IMU data will experience significant drift. The relatively high cost of DVLs also inhibits multirobot or swarm operations. In this article, we present the results of using fully connected feedforward neural networks to predict vehicle velocity (surge, sway, and heave) for AUV localization. Approximation functions for the vehicle velocities are derived based on the generalized six degree-of-freedom dynamic equation and kinematics, proving that they are suitable for feedforward neural networks with measurable inputs. The neural networks are trained using field data collected using an AUV. An ablation study was conducted to investigate the impacts of hyperparameters, and the most effective models were selected using the Akaike information criterion. The selected neural network models are tested on three new sets of mission data collected on two different days. The quantitative results show that localization using vehicle velocities predicted by neural networks yields lower averaged root-mean-square distance errors and standard deviation compared to the DVL-aided dead-reckoning method. However, the extent of this improvement was found to depend on ocean current conditions. The promising results show that lightweight fully connected feedforward neural networks offer an attractive approach for inertial-based underwater localization when DVLs or their measurements are not available.
\end{abstract}


\setstretch{1.25}
\IEEEpeerreviewmaketitle

\section{Introduction}\label{sec:I}

AUTONOMOUS underwater vehicles (AUVs) have been widely used for ocean observation and monitoring. Georeferencing the observations collected by an AUV is critical to produce meaningful measurements (such as seabed images, current profiles, and dissolved oxygen observations) for quantifying the spatial extents of ocean processes. However, positioning AUVs underwater is challenging due to the lack of GPS signals underwater. Over the years, AUV localization has been studied extensively \cite{Paull2014AUVNA}, with methods that can be typically divided into two types: external-aided methods and self-contained methods. 

External-aided approaches usually leverage acoustic devices, such as long baseline (LBL), short baseline (SBL), and ultra-SBL (USBL). More recently, two-way-travel-time and one-way-travel-time localization methods have been extensively studied \cite{Harris2021CooperativeAN}, \cite{Iv2019ANS}, \cite{Fischell2019RelativeAA}, \cite{Rypkema2018ClosedLoopSP}, \cite{Claus2018ClosedLoop}, \cite{Webster2012Advances}. These methods use the time-of-flight of acoustic messages to estimate the relative range between an underwater vehicle and another device at a known location. The range measurement is then fused using an extended Kalman filter (EKF) to update the vehicle pose. Acoustic-based methods work well in a bounded area, typically within several kilometers, where an AUV is within the reception range of the acoustic instrument. In contrast, long-range acoustic-aided localization methods are rarely studied \cite{Webster2015Towards} as the sound speed variations in the water columns increase the measurement uncertainty. Furthermore, the update rate when using acoustic communication devices is relatively slow, typically several seconds or even longer at further distances. Therefore, a self-contained localization method is still needed to estimate the vehicle position between acoustic receptions. The acoustic-aided localization also complicates the logistical operations, as acoustic instruments have to be deployed and recovered before and after the missions. 

Given these limitations, self-contained localization approaches emerge as favorable alternatives because they utilize sensors onboard the vehicle to estimate vehicle positions. Compared to external-aided approaches, it is appealing due to its simplicity of setup and ease of use. Dead-reckoning, also referred to as inertial-based navigation, is the most common self-contained method. The simplest dead-reckoning can be implemented using a kinematic model where the vehicle body frame velocity is estimated based on actuator commands (e.g., the duty cycle, r/min, current, or voltage of the thrusters) \cite{Claus2015TerrainaidedNF}. The velocity is then transformed into the Earth frame based on the orientation estimated from an inertial measurement unit (IMU) and integrated from a known location (e.g., the last GPS fix before being submerged) to estimate the underwater locations. However, due to the noise in the IMU measurements, uncertainty in thruster-velocity mapping, and ocean currents, the error in dead-reckoning will accumulate with a drift up to several kilometer per hour \cite{SupunAT2018ImplementationOA}. To help reduce such errors, a Doppler velocity log (DVL) is typically installed on a modern AUV to measure the vehicle's velocity over the ground. The DVL-aided dead-reckoning algorithm normally uses a filter (e.g., EKF) to merge DVL and IMU measurements, and the drift rate can be reduced to about 5\% or lower \cite{Zhao2022TowardsUS}. Over the years, DVL costs have decreased, but a decent DVL still costs at least several thousand US dollars. Besides the cost, DVL-aided dead-reckoning requires the AUV to maintain consistent DVL bottom locks, meaning it must operate within the DVL's range off the seabed. This limitation restricts the AUV operation near the seabed, making the AUV less applicable for mid-water surveys to address some important scientific questions \cite{Yoerger2021AHU}. 

Terrain-aided navigation (TAN) is another method that has gained attention in recent years (see \cite{Claus2015TerrainaidedNF}, \cite{Salavasidis2021TerrainAidedNW}, and \cite{Salavasidis2018TerrainaidedNF}) because it is self-contained and provides bounded error. Using TAN requires the AUV to stay within the sonar range off the seabed and mandates access to a prior map of the seafloor. For achieving a satisfactory TAN result, variations in seafloor elevation are desired for bounding the uncertainty growth \cite{Zhou2019WorkingTA}. Perception-aided simultaneous localization and mapping (SLAM) has recently gained significant traction because it eliminates the requirement for prior environmental knowledge, i.e., the terrain map \cite{Hidalgo2015ReviewOU}. At its core, the method leverages features detected by onboard perception sensors (e.g., cameras and sonars) at different moments to constrain the drift in dead reckoning \cite{Zhao2023TightlyCoupledVD}. Underwater SLAM has experienced slower progress compared to land-based SLAM, primarily due to challenges, such as sensor noise and reduced features. Nevertheless, seminal works have been done in underwater SLAM using multibeam sonar point clouds (see \cite{Barkby2011BathymetricSW} and \cite{Krasnosky2022AMP}), fusing visual and sonar measurements (see \cite{Zhao2023TightlyCoupledVD},\cite{Rahman2018SVIn2AU}, and \cite{Rahman2018SonarVI}), with multi-AUVs \cite{Paull2015CommunicationconstrainedMC}, and imaging sonars (see \cite{Suresh2020ActiveSU} and \cite{Li2018PoseGraphSU}). The main disadvantage of SLAM is its computational intensity since nonlinear optimization is needed to estimate the feature locations. As a consequence, the overall vehicle endurance may be reduced due to the need for a powerful computer. Furthermore, the robustness of front-end feature tracking is not universally applicable, as different scenes may require different imaging enhancement, feature detection and tracking algorithms to prevent failures in finding corresponding features observed at different times. 

In this article, our research is focused on the self-contained localization methods when there is no DVL measurement available or the AUV is not near the seabed. Under the above conditions, model-aided approaches have been explored where a vehicle dynamics model is built based on prior data to provide velocity estimations for vehicle localization using IMU measurements. Even though the model-aided approach will suffer from drift, it has shown promising improvement (accuracy and consistency) when compared to DVL-aided approaches. For example, the model-aided approach presented in \cite{Arnold2018RobustMI} could fuse the estimation from a hydrodynamic model and inertial measurements for localization. The method also accepts additional information about the relative velocity between the vehicle and the water measured by an acoustic Doppler current profiler (ADCP) to further improve the performance. Overall, the drift is about 75 m/h with a relatively expensive IMU (over \$10 k). In a recent work \cite{SupunAT2018ImplementationOA}, a simplified hydrodynamic model was proposed to estimate velocity in the Earth frame. The model coefficients were estimated using the prediction error method from prior field trial data where the vehicle velocities were estimated by differentiating the vehicle position measured by an LBL system. In \cite{SupunAT2018ImplementationOA}, the model uses the IMU-based forward speed from the front seat computer and angular acceleration by differentiating the gyro. However, this approach is less applicable to other vehicles, as the IMU-based forward speed may vary across different AUVs, and the angular acceleration is rate dependent. When an IMU is operated at a high sampling rate, such as 100 Hz, significant noise and erroneous acceleration and angular velocity measurements can be observed. Nevertheless, promising results were obtained from field tests, showing that the maximum drift is about 100 m/h when predicting the velocity using an affordable IMU (less than \$1 k). As discussed in this article, the localization performance is also environment dependent, as the impact of the water current is not included in the model. Due to the fact that the AUV model is highly nonlinear, the multivariable regression used in many model identification studies \cite{SupunAT2018ImplementationOA} may not provide a good model approximation and have uncertainties when predicting the vehicle states for localization. To this end, some advanced approaches, such as unsupervised online system identification, were explored (see \cite{Karras2019UnsupervisedOS} and \cite{Karras2013OnlineIO}). 

Recently, machine learning techniques, such as neural networks (NN) and deep learning, have been explored for underwater localization. Zhang et al. \cite{Zhang2020NavnetAN} introduced NavNet, a deep sequential learning-based navigation framework. It uses a recurrent neural network (RNN) and an attention mechanism to approximate the system model and measurement model for underwater localization. Improvements were shown when compared to conventional EKF and unscented Kalman filter sensor fusion methods. However, the proposed framework still needs DVL measurements as an input during operation. To the best of the authors' knowledge, the research in DVL-free localization strategies using deep learning is still limited. A recent study \cite{Topini2023AnEC} has investigated five different deep learning methods for predicting AUV's surge speed based on the navigation data collected with AutoSub500 and AutoSubLR. In \cite{Saksvik2021ADL}, an RNN was trained to predict vehicle surge and sway velocity. The paper demonstrated the approach on the Tethys AUV using field trial data and on a simulated underwater glider. A good agreement between the vehicle track estimated from the RNN outputs and DVL outputs was found. 

This article takes a similar approach presented in \cite{Saksvik2021ADL} where NN models are trained to predict body-frame velocities. When training the NN models, a DVL is needed to provide reference velocity measurements. Once the models are trained, the NN models will predict the vehicle's linear velocity components for localization, allowing the AUV to localize itself without the DVL or if it is away from the seabed beyond the DVL range. In the scenario that the DVL has been removed after the training for the purpose of reducing the overall AUV cost, a ``dummy'' DVL should be placed at the same location as the DVL. The dummy DVL should have a similar in-water weight and shape such that the vehicle's kinematics and dynamic characteristics remain unchanged. The dummy DVL could be 3-D printed to have the same shape as the actual DVL and have small voids for adding small metal inserts (e.g., washers and nuts) for fine-tuning its in-water weight to match the real DVL used when collecting the training data. 

Compared to the existing works (particularly, \cite{Saksvik2021ADL}), the work presented in this article has three major differences. First, we provide a detailed explanation of deriving the approximation function of the vehicle velocity based on the six degree-of-freedom marine vehicle dynamic equation, justifying our NN design (inputs and feasibility). Second, the feedforward neural network (FFN) is used, which is relatively small (less than 1500 parameters for the most complicated model) and fast to train (an hour per training session) compared to the ones used in the existing works. Third, we have added system voltage and current draw as model inputs, which will affect the forces generated from the thrusters that will ultimately impact vehicle velocity components. To the best of the authors' knowledge, research on the integration of FFN for underwater inertial-based localization is still rare, and this article seeks to advance this topic with the following contributions. \begin{enumerate}
\item This article has derived the approximation function of the vehicle linear velocities (see Section~\ref{sec:II}) based on the general 6-DOF marine vehicle dynamic equation of motion. We show that the linear velocities in the vehicle body frame are nonlinear functions of measurable vehicle states, including linear acceleration, angular velocity, orientation, thruster commands, system voltage and current, and vehicle depth. Three FFN are trained separately for predicting individual vehicle velocities along three body frame axes. The separation allows us to avoid the influence due to the skewness in the data. For example, for an AUV performing survey missions, its sway and heave motion are relatively small compared to the surge velocities. 
\item An ablation study (see Section~\ref{sec:IV}) is conducted to investigate the effects of hyperparameters (including batch size, neuron numbers, and layer numbers). We found that two layers with eight neurons are effective for these tasks, indicated by the minimum Akaike information criterion (AIC) values. The network is significantly smaller (less than 500 parameters) compared to relevant works (see \cite{Topini2023AnEC} and \cite{Saksvik2021ADL}) where complicated NNs with a large amount of parameters (over 1000) are used. This finding makes the method presented in this article applicable to computing resource-limited AUVs. 
\item We have validated the performance of localization using the velocity predicted from the trained NN models. We found that the new method has improved performance when compared to DVL-aided methods, especially when there is a relatively strong ocean current. The method also produces a similar performance compared to the model-aided DVL-free approaches (see \cite{SupunAT2018ImplementationOA} and \cite{Arnold2018RobustMI}) with a less expensive instrument setup.
\end{enumerate}

\section{NN Models Design}\label{sec:II}

\begin{figure}[htbp]
    \centering
    \includegraphics[width=0.95\textwidth]{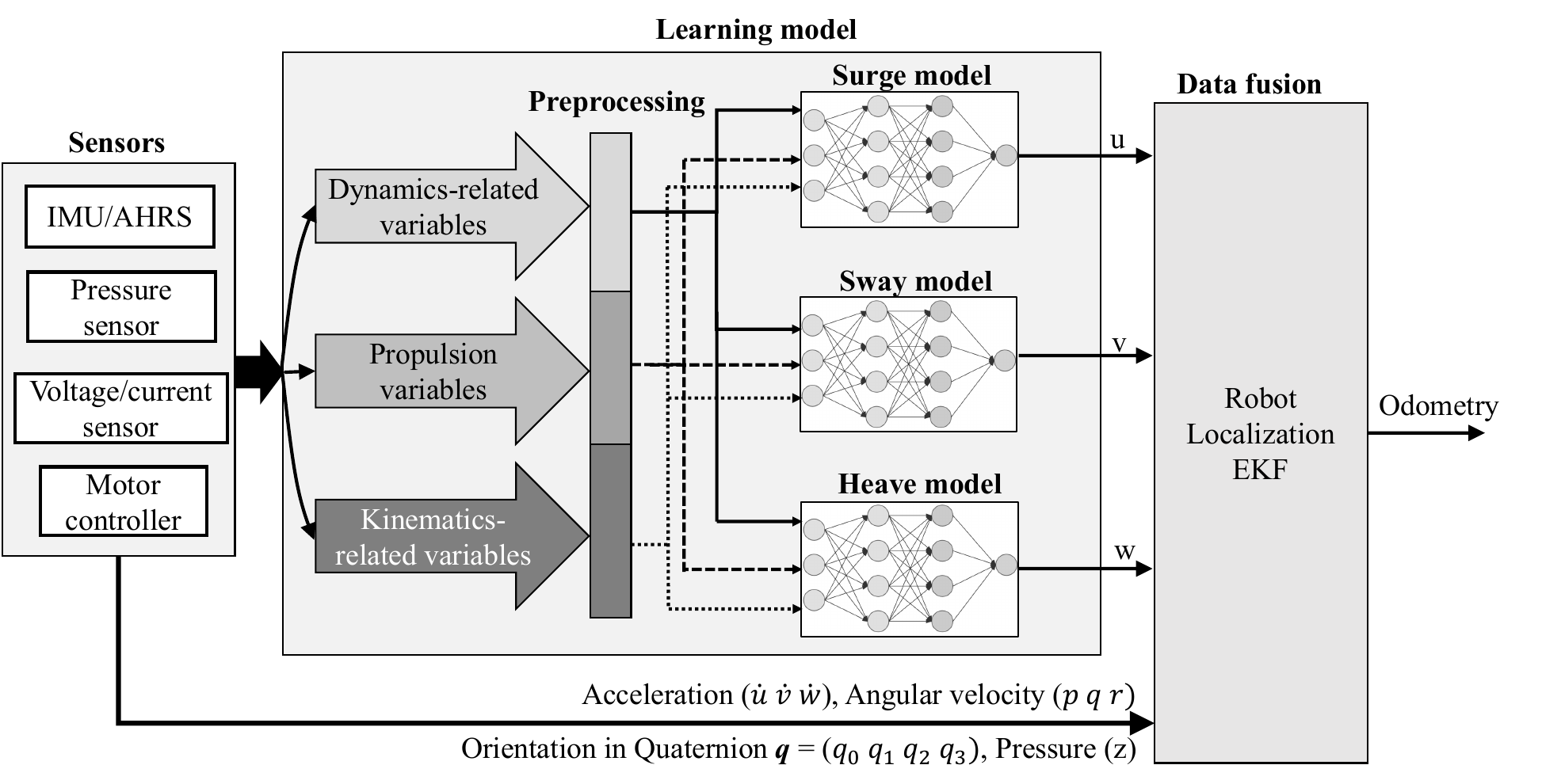}
    \caption{Overall system diagram of integrating sensor data and learning models for localization. The inputs of the models include four groups of measurable data: accelerations ($\dot{u}$, $\dot{v}$, and $\dot{w}$), angular velocity ($p$, $q$, and $r$), orientation in quaternion form ($q_0$, $q_1$, $q_2$, and $q_3$) measured by an IMU or AHRS; vehicle depth ($z$) measured by a pressure sensor; system voltage and current that can be measured using integrated chips; and motor commands from motor drivers. The three independent FFN use the data to predict vehicle body frame linear velocity $u$, $v$, and $w$, which is fused with IMU measurements and pressure sensor measurements to generate vehicle odometry.}
    \label{fig:system_diagram}
\end{figure}

The design of the NN model, regarding the proper choice of inputs, outputs, and preprocessing methods, is an important factor in achieving a satisfactory regression result. Since an AUV's surge, sway, and heave motion will be in different ranges, we used three deep neural network models to predict individual velocity components. Such separation of degree of freedoms is also recommended in \cite{Ven2004NeuralNA} when estimating hydrodynamic coefficients. As shown in Fig.~\ref{fig:system_diagram}, the inputs for each NN can be divided into three groups: dynamic-related inputs, propulsion variables, and kinematics-related inputs. These inputs are measurable using common sensors, such as IMU, attitude heading reference system (AHRS), pressure sensor, voltage/current sensor, and pulsewidth modulation (PWM) signals commanded to the motor controller. The outputs of the NN models are vehicle velocities, which can be measured by the DVL for model training. 

The dynamic-related inputs are selected based on the 6-DOF dynamic equation of motion of a marine vehicle, as shown in \eqref{eq:std_6dof}, derived from \cite{Fossen1994GuidanceAC} based on SNAME 1950 notation. On the left-hand side of \eqref{eq:std_6dof}, the force and torque induced by the propulsion system, environment, and gravity and buoyancy are denoted as $\boldsymbol{\tau}_p$, $\boldsymbol{\tau}_e$, and $\boldsymbol{\tau}_g$, respectively. On the right-hand side of \eqref{eq:std_6dof}, the equation includes the rigid-body mass and inertia ($\boldsymbol{M}_{RB}$), added mass matrix ($\boldsymbol{M}_A$), Coriolis/centripetal terms due to vehicle mass and inertia ($\boldsymbol{C}_{RB}$) and added mass ($\boldsymbol{C}_A$), and hydrodynamic damping force and torque ($\boldsymbol{D}$). The velocity, $\boldsymbol{\nu}$, is defined in \eqref{eq:velocity}, where the linear velocity ($\boldsymbol{\nu}_1$) is the output of the NN model, and the angular velocity ($\boldsymbol{\nu}_2$) can be measured by a gyroscope in an IMU

\begin{equation}
\label{eq:std_6dof}
    \boldsymbol{\tau}_p + \boldsymbol{\tau}_e + \boldsymbol{\tau}_g = (\boldsymbol{M}_{RB} + \boldsymbol{M}_{A}) \dot{\boldsymbol{\nu}} + (\boldsymbol{C}_{RB}(\boldsymbol{\nu})+\boldsymbol{C}_A (\boldsymbol{\nu})) \boldsymbol{\nu} + \boldsymbol{D}(\boldsymbol{\nu}) \boldsymbol{v}
\end{equation}

\begin{equation}
\label{eq:velocity}
 \boldsymbol{\nu} = [\boldsymbol{\nu}_1^\top, \boldsymbol{\nu}_2^\top]
 \quad \text{where}\quad
 \boldsymbol{\nu}_1= [u, v, w]^\top
 \boldsymbol{\nu}_2 = [p, q, r]^\top
\end{equation}

\begin{subequations}
    \label{eq:expanded_velocities}
\begin{align}
\label{eq:expanded_u}
\begin{array}{cl}
     X=&  m \dot{u} + m z_G \dot{q} - m y_G \dot{r} \\
     & + m(y_G q + z_G r) p - m(x_G q - w) q - m(x_G r + v) r\\
     & + X_{\dot{u}} \dot{u} - Z_{\dot{w}} w q + Y_{\dot{v}} v r + X_{u^2}|u|u\\
\end{array} \\
\label{eq:expanded_v}
\begin{array}{cl}
    Y=&  m \dot{v} - m z_G \dot{p} - m x_G \dot{r} \\
     & - m(y_G p + w) p + m(z_G r - x_G p) q - m(y_G r + u) r\\
     & + Y_{\dot{v}} \dot{v} + Z_{\dot{w}} w p + X_{\dot{u}} u r + Y_{v^2}|v|v\\
\end{array} \\
\label{eq:expanded_w}
\begin{array}{cl}
    Z=&  m \dot{w} + m y_G \dot{p} - m x_G \dot{r} \\
     & - m(z_G p - v) p - m(z_G q + u) q + m(x_G p + y_G q) r\\
     & + Z_{\dot{w}} \dot{w} - Y_{\dot{v}} v p + X_{\dot{u}} u q + Z_{w^2}|w|w\\
\end{array}
\end{align}
\end{subequations}

Next, we expand \eqref{eq:std_6dof} for surge, sway, and heave into \eqref{eq:expanded_u}-- \eqref{eq:expanded_w}, where $X$, $Y$, and $Z$ on the left-hand side represent the overall forces induced by the propulsion system, environment, and restoring forces in the vehicle's body frame. On the right-hand side, $m$ is the vehicle mass, $x_G$, $y_G$, and $z_G$ indicate the location of the center of gravity, $X_{\dot{u}}$, $Y_{\dot{v}}$, and $Z_{\dot{w}}$ are the added mass coefficients, and $X_{u^2}$, $Y_{v^2}$, and $Z_{w^2}$ are hydrodynamic damping coefficients. 

We make two additional assumptions to simplify the equations. First, we assume that the origin of the body frame is located at the center of gravity such that the terms with $x_G$, $y_G$, and $z_G$ become zeros. Second, we remove the coupled terms involving rotation and translation [such as $wq$ in \eqref{eq:expanded_velocities}] because they are relatively small compared to added mass and hydrodynamic damping terms. After removing the negligible terms from the above assumptions, \eqref{eq:expanded_velocities} is reorganized into \eqref{eq:expanded2}, which is a generic equation typically used for system identification \cite{Li2014ModellingAI} for surge, sway, and heave motion with lump-sum coefficients denoted as $A_{ij}$. It is important to note that linear acceleration terms now only appear in their corresponding axis on the right-hand side of \eqref{eq:expanded2}. This later allows us to provide different acceleration inputs for the three NN models. The left-hand side of \eqref{eq:expanded2} consists of three components, restoring force ($\boldsymbol{F}_g$), propulsion force ($\boldsymbol{F}_p$), and environmental force ($\boldsymbol{F}_e$). The restoring force depends on the vehicle's orientation ($\boldsymbol{q}$) represented using a quaternion and the submerged volume (related to depth $z^e$). The propulsion force generated by thrusters depends on the thruster commands ($\boldsymbol{u}_T$), vehicle velocity ($\boldsymbol{\nu}$), system voltage ($V$), and current draw ($I$). For the ith thruster, its thrust ($F_p^i$) can be formulated using the thruster model shown in \eqref{eq:thrust_eq} \cite{Li2014ModellingAI}, where $u_T^i$ is the thruster command, $A_{T1}$ and $A_{T2}$ are the thruster coefficients that are depended on the supply voltage ($V$), $\nu_a^i$ is the advance velocity that the ith thruster is experiencing along its x-axis. The advance velocity of a thruster can be further expressed in \eqref{eq:advance_velocity} as a function of vehicle velocity ($\boldsymbol{\nu}_1$ and $\boldsymbol{\nu}_2$), thruster location ($\boldsymbol{r}_i^b$) in the body frame, and rotation matrix ($\boldsymbol{R}_b^i$) from body frame to the ith thruster frame. Meanwhile, current draw ($I$) is also an indicator of thrust as it relates to motor shaft torque \cite{Claus2010DevelopmentOA}.

The environmental force $\boldsymbol{F}_e$ depends on the relative velocity between the vehicle and the current, i.e., the velocity of the current expressed in the body frame ($\boldsymbol{\nu}_w^b$). However, we have excluded the relative velocity in the NN model since an ADCP or a DVL with water-track capability will be required to measure the relative motion

\begin{gather}
    \boldsymbol{F_g}(\boldsymbol{q}, z^e) + \boldsymbol{F_p}(\boldsymbol{u}_T,\boldsymbol{\nu}, V, I)+ \boldsymbol{F_e}(\boldsymbol{\nu}^b_w)
    =
    \left[
    \begin{array}{l}
         A_{11} \dot{u} + A_{12}|u|u \\
         A_{21} \dot{v} + A_{22}|v|v \\
         A_{31} \dot{w} + A_{32}|w|w  \\
    \end{array}\right]. \label{eq:expanded2} \\
    F_p^i = A_{T1}(V) \nu_a^i u_T + A_{T2}(V) |u_T^i| u_T^i \label{eq:thrust_eq} \\
    \nu^i_a =  
    \begin{array}{ccc}
    [1&0&0]
    \end{array}
    \boldsymbol{R}^i_b (\boldsymbol{\nu}_1 + \boldsymbol{r}^b_i \times \boldsymbol{\nu}_2).
    \label{eq:advance_velocity}
\end{gather}

Next, we derive two kinematic relationships to show that the linear velocities $\boldsymbol{\nu}_1$ are related to angular velocity and depth rates, which are included as NN inputs. In \eqref{eq:v_fun_1}, we present the relationship between the vehicle velocity ($\boldsymbol{\nu}_1$ and $\boldsymbol{\nu}_2$) in the body frame and the DVL measured linear velocity ($\boldsymbol{\nu}_1^d$). Since the DVL may be offset with a translation $\boldsymbol{T}_d^b$ from the origin of the body frame (typically the location of the IMU), the DVL measurements are also affected by the angular velocity ($\boldsymbol{\nu}_2$) of the vehicle body. Therefore, when the DVL measurements are used for training an NN model to predict the vehicle's linear velocity, we need to include angular velocities, but these terms are already included as dynamic-related inputs, as shown in \eqref{eq:expanded2}. Furthermore, in \eqref{eq:kinematics}, we show that the depth rate $\dot{z}^e$ (measurable using a pressure sensor) is related to the body-frame linear velocity ($\boldsymbol{\nu}_1$) through the rotation matrix from the body frame to Earth frame ($\boldsymbol{R}_b^e$), which is constructed using the vehicle orientation $\boldsymbol{q}$

\begin{equation}
    \label{eq:v_fun_1}
    \boldsymbol{\nu}_1^{d} = \boldsymbol{R}^d_b (\boldsymbol{\nu}_1 + \boldsymbol{\nu}_2 \times \boldsymbol{T}^{b}_{d})
\end{equation}

\begin{equation}
    \label{eq:kinematics}
    \left[
    \dot{x}^e,
    \dot{y}^e,
    \dot{z}^e
    \right]^\top
    = \boldsymbol{R}^e_b(\boldsymbol{q}) 
   \boldsymbol{\nu}_1
\end{equation}

\begin{subequations}
    \label{eq:final_inputs}
\begin{align}
     u  &= f_u(\dot{u}, \boldsymbol{\nu}_2, \boldsymbol{q}, \boldsymbol{u}_T, V, I, z^e,\dot{z}^e)\\
     v  &= f_v(\dot{v}, \boldsymbol{\nu}_2, \boldsymbol{q}, \boldsymbol{u}_T, V, I, z^e,\dot{z}^e)\\
     w  &= f_w(\dot{w}, \boldsymbol{\nu}_2, \boldsymbol{q}, \boldsymbol{u}_T, V, I, z^e, \dot{z}^e).
\end{align}
\end{subequations}

\begin{table}[htbp]
    \caption{Summary of the inputs of the NN for predicting body frame linear velocity}
    \centering
    \begin{tabular}{c|c|c|c|c|c|c|c|c|c|c}
    \hline
    &\multicolumn{3}{c|}{linear accel.}& ang. vel.&orientation&thrusters&\multicolumn{2}{c|}{Power}&\multicolumn{2}{c}{depth}\\
    \hline
    Prediction&$\dot{u}$&$\dot{v}$&$\dot{w}$&$p,q,r$&$\boldsymbol{q}$&$\boldsymbol{u}_T $ &V&I&$z^e$&$\dot{z}^e$\\
    \hline
    $u$ network&Y&N&N&Y&Y&Y&Y&Y&Y&Y\\
    \hline
    $v$ network&N&Y&N&Y&Y&Y&Y&Y&Y&Y\\
    \hline
    $w$ network&N&N&Y&Y&Y&Y&Y&Y&Y&Y\\
    \hline
    \end{tabular}
    \label{tab:inputs}
\end{table}

Having analyzed the contributions to the linear velocities from dynamics, \eqref{eq:expanded2}--\eqref{eq:advance_velocity}, and kinematics, \eqref{eq:v_fun_1} and \eqref{eq:kinematics}, we see that the linear velocity of the vehicle can be indirectly observed from these relationships, which are highly nonlinear. Therefore, we combine the relationships by expressing the vehicle body-frame velocity components as nonlinear functions of the measurable inputs, as shown in \eqref{eq:final_inputs}. We suggest to construct mapping functions for surge, sway, and heave velocity separately and use NN models to approximate them. The inputs of the NN models consist of linear acceleration in corresponding axis, angular velocity ($\boldsymbol{\nu}_2$), quaternion orientation ($\boldsymbol{q}$), thruster commands ($\boldsymbol{u}_T$, four values in our case), system voltage ($V$), system current drain ($I$), depth ($z^e$), and depth rate ($\dot{z}^e$), as summarized in Table~\ref{tab:inputs}. As mentioned earlier, all the inputs are measurable using sensors, making our NN models realistic to implement on actual AUVs. Noting that, similar to system identification, the training of the NN models is vehicle-specific due to the differences in hydrodynamics and physics.

\section{Field Data Collection}\label{sec:III}

\begin{figure}[htbp]
    \centering
    \includegraphics[width=0.7\textwidth]{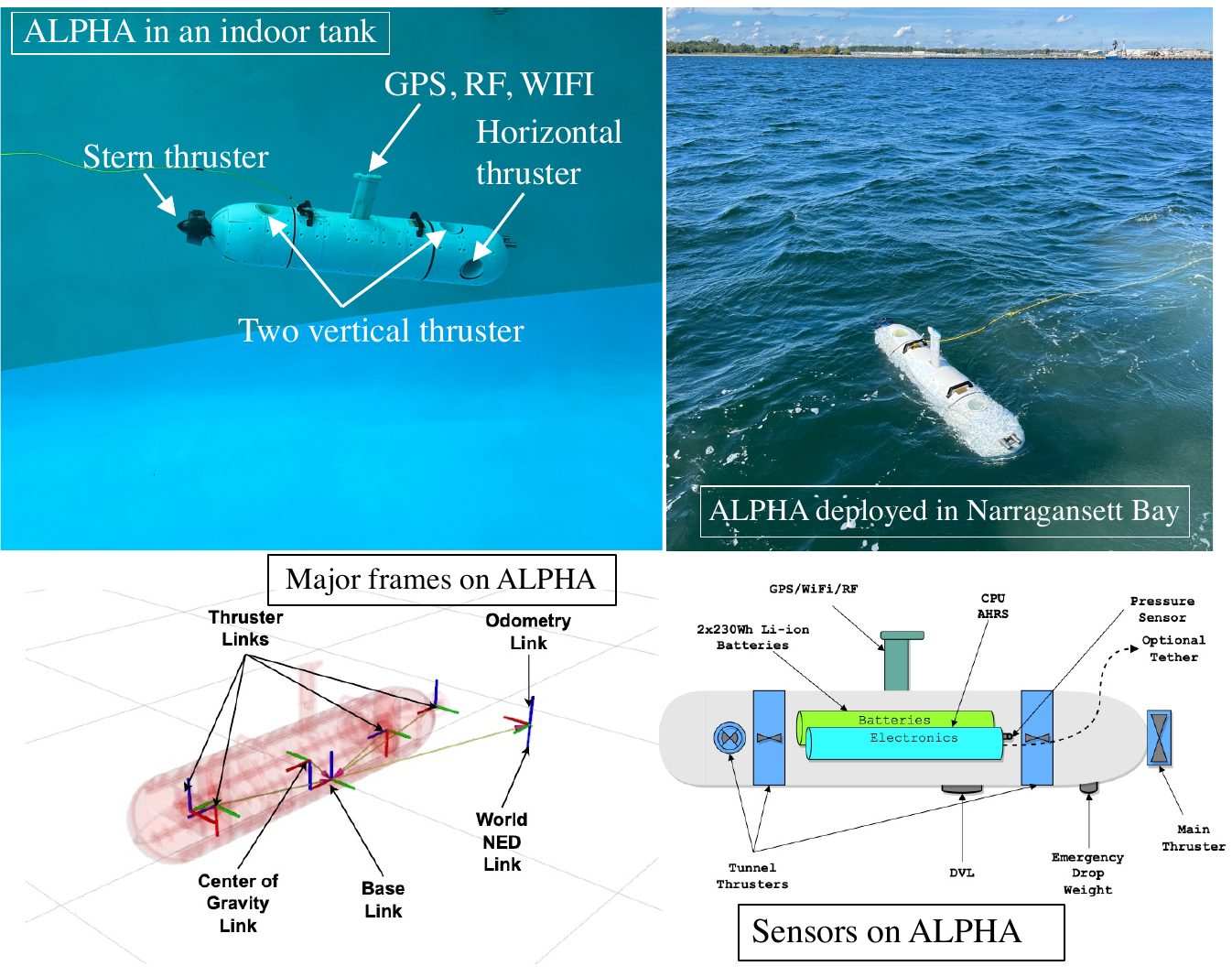}
    \caption{ALPHA deployed in (a) an indoor water tank and (b) Narragansett Bay. Major frames on AUV are illustrated in (c) with major components shown in (d).}
    \label{fig:alpha}
\end{figure}

We have conducted several field trials in Narragansett Bay with ALPHA \cite{Zhou2022AcrobaticLP}, a highly maneuverable underwater vehicle equipped with one stern thruster, two vertical tunnel thrusters, and a horizontal thruster, as shown in Fig.~\ref{fig:alpha}. A suite of sensors is available on the AUV to collect the needed data for training the NN models. Specifically, the Xsens MTi-630 IMU was used to collect motion measurements (including acceleration, angular velocity, and vehicle orientation) at 100 Hz. A pressure sensor (Bar30 from Blue Robotics) is used to collect depth measurements and to derive the depth rate. The thruster commands (range between $-$1 and 1) are updated at 10 Hz by the vehicle's low-level proportional--integral--derivative pose controller. The commands are converted into PWM signals by a microcontroller and transmitted to the motor controllers. A power monitoring breakout board is installed to record the battery voltage and the current draw of the system. Finally, a DVL (Waterlinked A50) is running at 5 Hz to collect measurements of the linear velocity of the vehicle. 

\begin{figure}[htbp]
    \centering
    \includegraphics[width=1.0\textwidth]{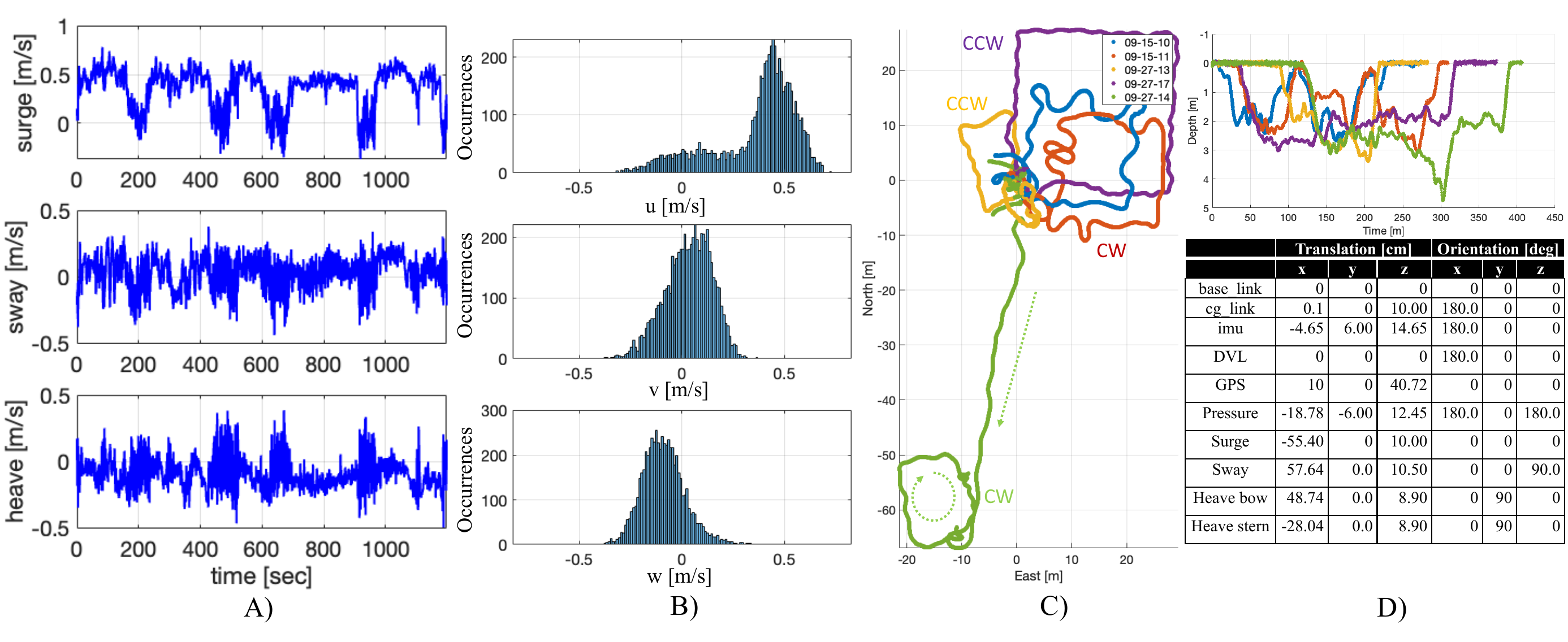}
    \caption{(a) DVL measured velocity in the training data. (b) Histogram of the DVL measured velocities in the training data. (c) Vehicle trajectories in the training data. (d) Vehicle depth in the training data sets, and the transformation between different local coordinate systems is listed in (e).}
    \label{fig:training_path}
\end{figure}

We have conducted a total of eight successful trials on two days. Among the eight deployments, we selected five of them for training the NN models with the vehicle trajectories shown in Fig.~\ref{fig:training_path}(c). The selected deployments are the ones with small environmental impacts where the surfacing locations indicated by the GPS fix and the position estimated by the localization system are very close (less than 5 m for a mission of about 15 min). The small errors suggest that the vehicle velocities measured in these trials are close to the actual speed-over-ground, which is beneficial for NN model training. As shown in Fig.~\ref{fig:training_path}, the missions for NN model training consist of different AUV movement patterns (a square shape pattern in both clockwise and counterclockwise directions, winding paths resulting from teleoperation, and a straight line path), which could capture sufficient data for training and generalizing the NN models. As shown in Fig.~\ref{fig:training_path}(a), the velocity varies during the trials, with the histogram shown in Fig.~\ref{fig:training_path}(b). The surge velocity has high occurrences around 0.5 m/s, which is the typical operation region of our AUV. 

The vehicle trajectories shown in Fig.~\ref{fig:training_path}(c) are calculated by fusing the measurements from IMU, DVL, and the pressure sensor using the EKF available from the robot localization \cite{MooreStouchKeneralizedEkf2014}. The EKF is developed based on the discrete time kinematic model (including relative position and orientation between sensors) without needing to know the vehicle dynamics. The same method is used to generate the benchmark dead-reckoning path in Section~\ref{sec:V}. 

The three remaining missions are excluded from training and solely used for testing the trained models in Section~\ref{sec:V}. In these missions, the AUV was driving to move in squared boxes, which is typical for testing guidance, navigation, and control systems and validating system identification results (see \cite{SupunAT2018ImplementationOA} and \cite{Arnold2018RobustMI}). In the test data set, we found the average ocean current is up to 5 cm/s (see Table~\ref{tab:current_correction} in Section~\ref{sec:V}) based on the comparison between the DVL-aided dead-reckoning track and the periodical surface GPS fixes. Among the three testing trials, we observed that the original odometry computed by the robot localization using the DVL, IMU, and pressure sensor measurements has moderate drift in two trials, and the third one has minimum drift (see detailed discussion in Section~\ref{sec:V}). Therefore, we could assess the consistency of the NN models under different ocean current impacts.

\section{Model Training and Selection}\label{sec:IV}

\textit{Preprocessing:} Since the data from IMU (100 Hz), DVL (5 Hz), low-level control (10 Hz), and power monitoring board (10 Hz) are sampled at different rates, we aligned the data based on the DVL's timestamp. The rationale is twofold. First, our NN models predict vehicle velocity that will replace the DVL measurements. Therefore, we want to avoid data manipulation (e.g., filtering, interpolation, and smoothing) on the DVL measurements, which may make the model less generalized, and the procedure becomes vehicle-specific. Second, the DVL has the lowest sampling rate. For the thruster commands, we used the previous neighbor interpolation, other than linear interpolation, since the thruster command will remain unchanged until a new command is available. In total, we have 7303 data points for training, equivalent to about 25 min of motion data. In the training data, we selected every fifth data point to form a validation data set (1462 data points) for model selection. Prior to the training, we rescaled all the inputs into a similar range to provide numerical consistency between the weights and biases in the NN models. Existing methods, such as min--max scaling, can also be used. However, the scaling ratio may be different if a different training data set is used, and this may be particularly challenging for continuous learning. For our case, the surge and sway acceleration ($\dot{u}$ and $\dot{v}$) are in the order of 10; therefore, we divided the values by 10. For the heave acceleration ($\dot{w}$), we offset the value by the gravity coefficient (9.8), then divide it by 10 to bring the measurements within $-$1 to 1. No changes are made to the angular velocity, quaternion, and depth rate since they are within the range of $-$1 to 1. We divided the voltage and current readings by 10 and 20, since the original range is between 12 and 15 V, and 1--25 A, respectively. Lastly, the model exclusively utilizes motion data collected when the vehicle is submerged for training. To perform data pruning, we applied a depth-based threshold of 0.6 m to only include the data when AUV is fully submerged. 

\begin{figure}[htbp]
    \centering
    \includegraphics[width=0.7\textwidth]{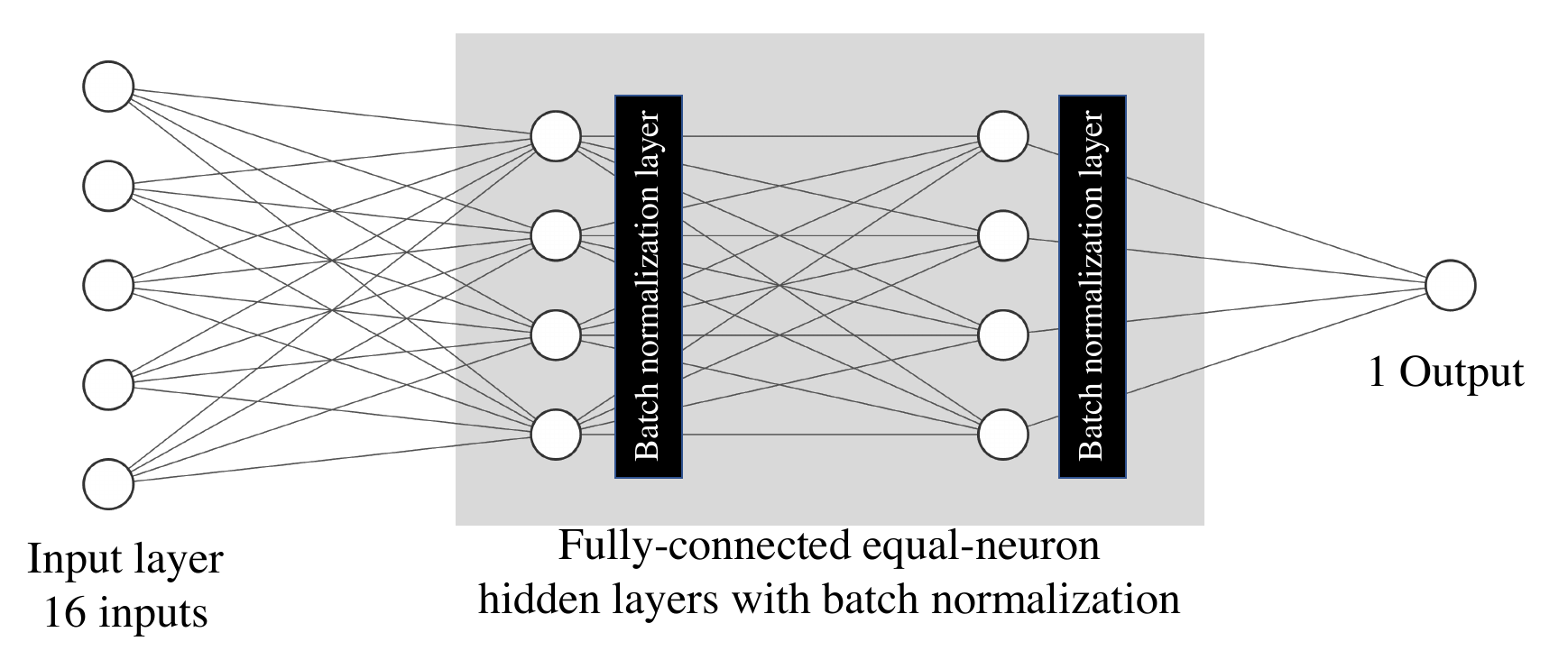}
    \caption{Example of our NN model with two hidden layers and four neurons in each layer.}
    \label{fig:NN_example}
\end{figure}

\begin{table}[htbp]
    \caption{List of adjustable and fixed model hyperparameters}
    \centering
    \begin{tabular}{c|c|c}
    \hline
         \multirow{3}{*}{Adjustable Hyperparameters}&Hidden layer number& {1, 2, 3, 4}  \\
         \cline{2-3}
         &Neuron number& {4, 8, 16}\\
         \cline{2-3}
         \cline{2-3}
         &Batch size & {8, 16, 32, 64, 128, 256}\\
         \cline{2-3}
         \cline{2-3}
         \hline
         \multirow{6}{*}{Fixed hyperparameters}&Learning rate & 0.005\\
         \cline{2-3}
         &Activation function & tanh\\
         \cline{2-3}
         &Regularization & Batch normalization\\
         \cline{2-3}
         &Optimizer & Adamax\\
         \cline{2-3}
         &Maximum epoch number & 20000\\
         \cline{2-3}
         &Early stopping & 100\\
         \hline
    \end{tabular}
    \label{tab:training_summary}
\end{table}

\textit{Hyperparameters:} There are many hyperparameters in an NN model, leading to different model complexities and performance. For example, the neuron number and hidden layer number affect the model size and complexity, while the batch size and regularization strategies impact the training time and convergence rate. To achieve the optimum network architecture, we have constructed different NN models with different hyperparameters. In this study, we employ a fully connected FFN (see Fig.~\ref{fig:NN_example}) to leverage its potential for online/continuous learning with a reduced number of parameters compared to RNNs. As listed in Table~\ref{tab:training_summary}, we have investigated three major hyperparameters. First, the hidden layer number greatly impacts the model's performance and complexity. In a general case, three hidden layers are sufficient to represent a highly nonlinear mapping from output and input \cite{NN_book}. Therefore, we tested hidden layer numbers ranging from 1 to 4, and each hidden layer has the same number of neurons. After each hidden layer, we used the tanh activation function to make sure the model is continuous and differentiable for possible theoretical analysis. For regularization, batch normalization \cite{Luo2018TowardsUR} is applied after the activation function. Moreover, we tested different numbers of neurons in exponentially increasing orders from 4 to 16 (capped at the number of inputs). Each model is trained with different batch sizes, increasing exponentially from 8 to 256. Meanwhile, some hyperparameters remain the same. For example, the learning rate is set to a constant (0.005), we used the batch normalization layer for regularization, and the Adamax optimizer \cite{Kingma2014AdamAM} is used in the training. The training stop condition remains the same for all models, with a maximum epoch number of 20 000 and early stopping of 100. The initial weights of the NN model are assigned using the Glorot uniform initialization \cite{Glorot2010UnderstandingTD} and the biases are initialized to zeros. The selected weight initialization will induce randomness in the model training. Therefore, we trained each model configuration 20 times. 

\begin{figure}[htbp]
    \centering
    \includegraphics[width=1.0\textwidth]{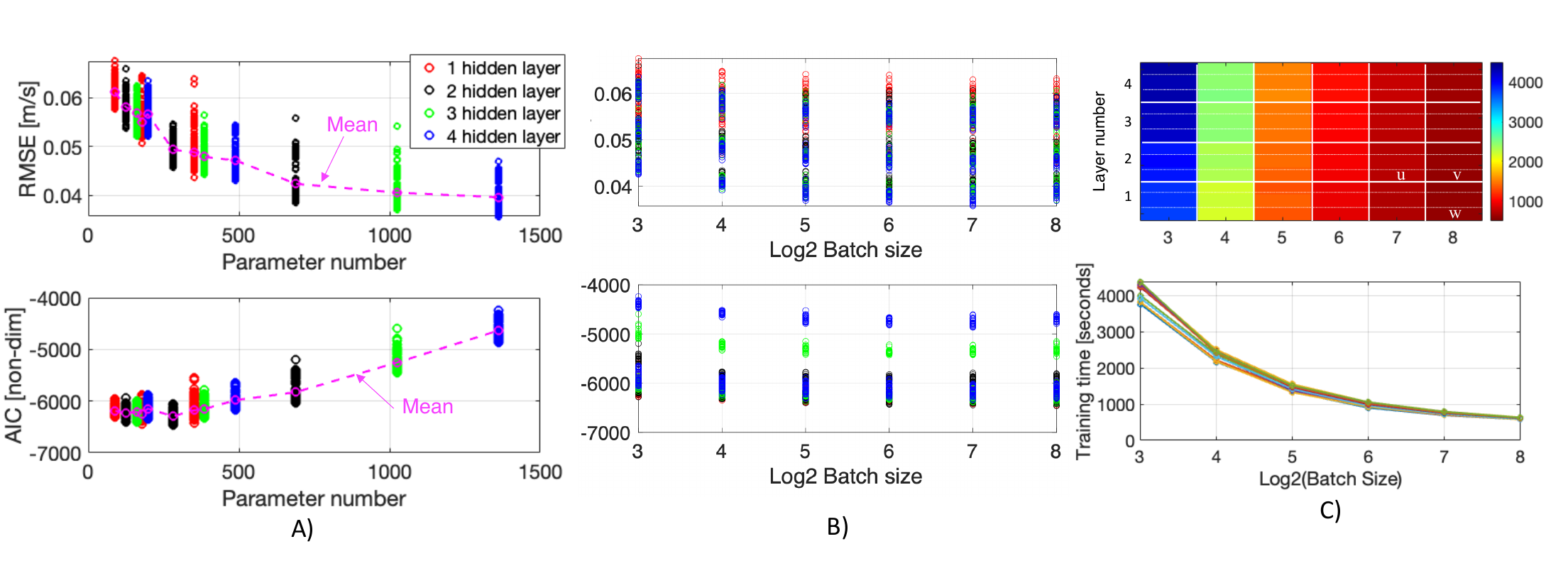}
    \caption{Hyperparameter impacts on RMSE, AIC values, and training time for estimating sway speed. (a) RMSE and AIC values change with respect to the total number of model parameters. (b) RMSE and AIC change with respect to batch size. (c) and (d) Training time (seconds) without early stopping with different layer numbers, neuron numbers, and batch sizes. For the same layer number and batch size separated by solid lines, the training time for the network with different neuron numbers (4, 8, and 16) is separated by dash-lines in each solid-line grid. The configuration of the selected models for surge, sway, and heave is also indicated in the heatmap.}
    \label{fig:validation_analysis}
\end{figure}

\textit{Model comparison and selection:} After training, we applied the validation data set to check the model performance. Herein, we are investigating two key performances, the root-mean-square error (RMSE) and the AIC value \cite{Akaike1969FittingAM}. The AIC value in \eqref{eq:AIC} is a metric commonly used for selecting an effective model, where it is a function of the number of data points in the validation data set ($n$), mean-squared error (inside the $\ln$), and the number of model parameters ($k$). For an effective model, the AIC value should be small since the mean-squared error and the number of parameters ($k$) should both be minimized. If two models produce the same mean-squared error, the model with fewer parameters will be deemed more effective, leading to a smaller AIC value. As shown in Fig.~\ref{fig:validation_analysis}(a), in general, we found that RMSE reduces with increasing number of model parameters, whereas AIC values are consistent when parameter number is less than 400 but increase rapidly after parameter number has exceeded 500. This may indicate that a model with the total parameter number less than 500 is sufficient to approximate our mapping functions shown in \eqref{eq:final_inputs}. Fig.~\ref{fig:validation_analysis}(a) and (b) presents the RMSE and AIC values for the sway models trained with different hyperparameters, where we observed the trend we discussed above. Moreover, we found that increasing the batch size has a small influence on reducing the AIC value and RMSE (especially for the models with three and four hidden layers, colored in blue and green in Fig.~\ref{fig:validation_analysis}(b). Finally, we show the training time (20 000 epochs without early stopping on an Apple M3 CPU) with different model setups in Fig.~\ref{fig:validation_analysis}(c) and (d). We observed that smaller batch sizes will cause exponential increases in overall training time. For model structure, layer number has a moderate impact on training time by comparing the heatmap in each column, whereas neuron number has minimum impact as the colors inside each solid-line grid are similar. Overall, the tested models are small (less than 1500 parameters with a memory usage of less than 1 MB) and could be deployed on modern single board computers (such as Raspberry Pi 4)

\begin{equation}
    \label{eq:AIC}
    AIC = n \ln{\frac{\sum_{i=1}^n{(y_i-\hat{y_i})^2}}{n}} + 2k.
\end{equation}

Among all the models, we selected the ones with the lowest AIC values for the surge, sway, and heave velocity prediction. Interestingly, we found that the selected models are trained using large batch sizes that could be trained in a relatively short time, less than 1000 s. 

\begin{table}[htbp]
    \caption{Hyperparameters for the selected models}
    \centering
    \begin{tabular}{c|c|c|c|c}
        \hline
         DOF&layer\#&neuron \#&model \#&batch size\\
         \hline
         u&2&8&14&128\\
         \hline
         v&2&8&16&256\\
         \hline
         w&1&8&11&256\\
         \hline
    \end{tabular}
    \label{tab:selected_models}
\end{table}

\begin{figure}[htbp]
    \centering
    \includegraphics[width=0.95\textwidth]{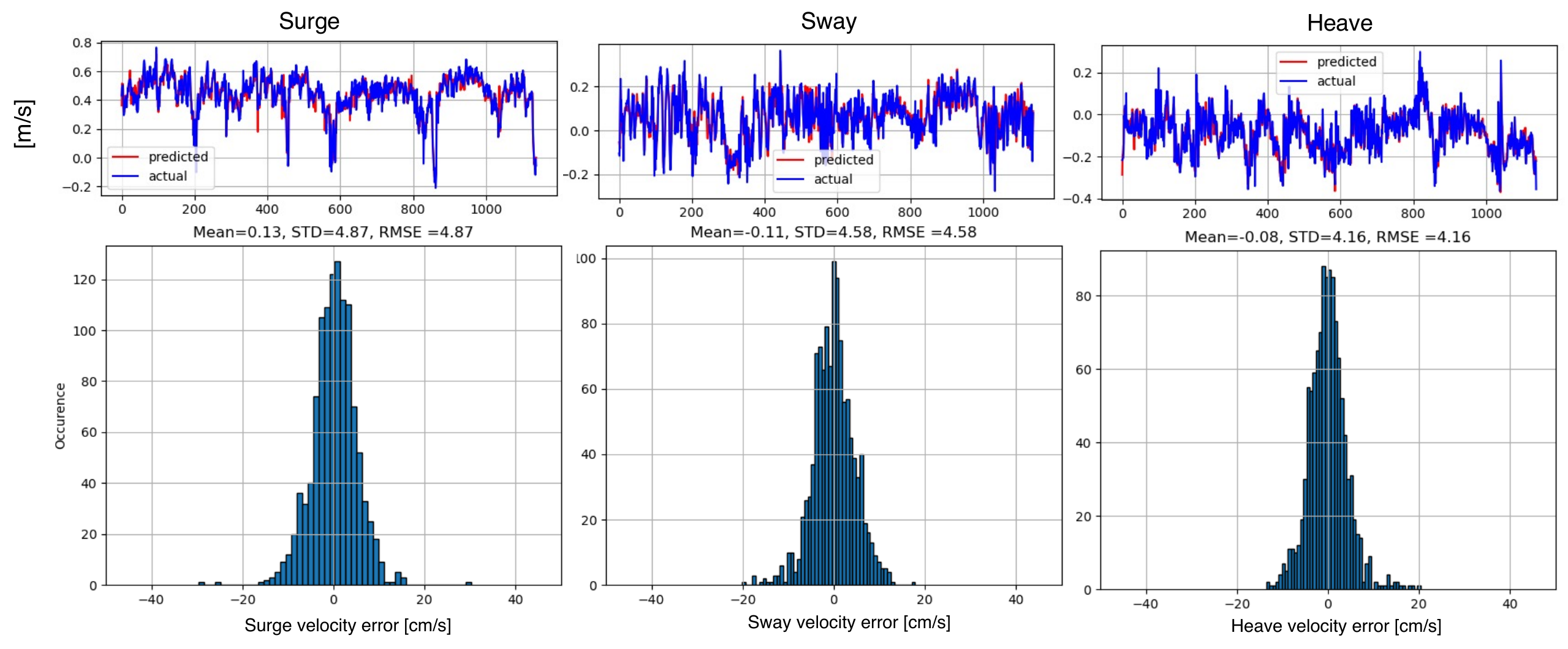}
    \caption{Comparison between the predicted and the actual values for the selected models using the validation data set.}
    \label{fig:validation}
\end{figure}

Table~\ref{tab:selected_models} lists the hyperparameters of the selected models. In Fig.~\ref{fig:validation}, we show the agreement between the outputs of the selected models and the actual values in the validation data set. RMSE values are observed at about 5 cm/s with small mean errors (about 0.1 cm/s) for surge, sway, and heave. Although the difference between the prediction and the actual value will still result in small drifts in odometry when fusing the NN model outputs with the IMU, the NN models have two major contributions to the AUV localization since DVL measurements will not become a requirement for computing odometry. First, it makes the localization more reliable since velocity estimates are still available when DVL is not providing measurements in cases of an AUV is away from the bottom or traveling over a soft bottom. Second, it opens the possibility for further decreasing the cost of an AUV for expandable applications and distributed deployments.

\section{Learning-Based Localization Results}\label{sec:V}

Herein, we combine the vehicle velocities predicted by the NN models and IMU measurements using EKF in the robot localization software to generate a new odometry. Then, it is compared with the original results (with DVL measurements) and a reference track (ocean current corrected). Essentially, we are replacing the original DVL measurements with the NN model outputs when computing the new vehicle path. The evaluation is done in playback mode with the three trials that were not used in model training and model selection. We run two localization setups simultaneously, where one is the original method that fuses the velocity from the DVL and the second one fuses the velocity predicted by the NN models. The initial position of the navigation tracks are defined at the first acceptable GPS fix (e.g., uncertainty less than 3 m). Therefore, all the tracks have the same initial position. Even though the NN model could provide velocity predictions at a much faster rate up to 90 Hz on a modern laptop (8-gen Intel i7), we program the NN model to produce velocity data at the same rate as the DVL measurements to make localization results comparable. 

As mentioned earlier, three test trials (15 September 2012, 15 September 2013, and 27 September 2017) were separated from the training and validation processes. For these trials, we corrected the impacts from ocean currents (as listed in Table~\ref{tab:current_correction}) by assuming a constant drift to align the odometry (based on DVL, IMU, and pressure sensor with EKF) with the locations of the GPS fixes. A similar approach has been used in the past to obtain the ``best'' underwater trajectory in postprocessing for AUVs and underwater gliders \cite{Zhou2017UnderwaterAN}. Worth noting that the GPS corrected reference path still contains uncertainty since the actual position of the vehicle when submerged is unknown, and the GPS sensor itself has uncertainty (accuracy of under 3 m 50\% circular error probable in our case). The values of the current correction used in each trial are listed in Table~\ref{tab:current_correction}. We can see that the current had a larger impact on the AUV's original odometry in the two trials conducted on 15 September. The corrected path is treated as the reference track for quantifying the localization performance when using the velocity predictions from the NN models. We also included the fourth trial (27 September 2016), where 80\% of the data from this trial were used in training, and 20\% were used for model selection. The fourth trial provides a benchmark (``best'') performance of the trained models. 

\begin{table}[htbp]
    \caption{General information for the field trials used for testing the NN models}
    \centering
    \begin{tabular}{c|c|c|c|c}
    \hline
    &\multicolumn{4}{c}{Trial number}\\
    \hline
    Values &09-15-12&09-15-13&09-27-17&09-27-16*\\
    \hline
    Current correction in x [cm/s]& -2.0 &-0.35&0.3&1.5\\
    \hline
    Current correction in y [cm/s]& 4.5&2.7&-1.2&-0.7\\
    \hline
    Total distance [m]&285.3&237.1&140.4&139.6\\
    \hline
    Mission time [min]&9.90&8.55&6.01&6.23\\
    \hline
    \end{tabular}

\smallskip
\footnotesize Note: the data from 27 September 2016 were used in model training and validation.
    \label{tab:current_correction}
\end{table}

\begin{figure}[htbp]
    \centering
    \includegraphics[width=0.6\textwidth]{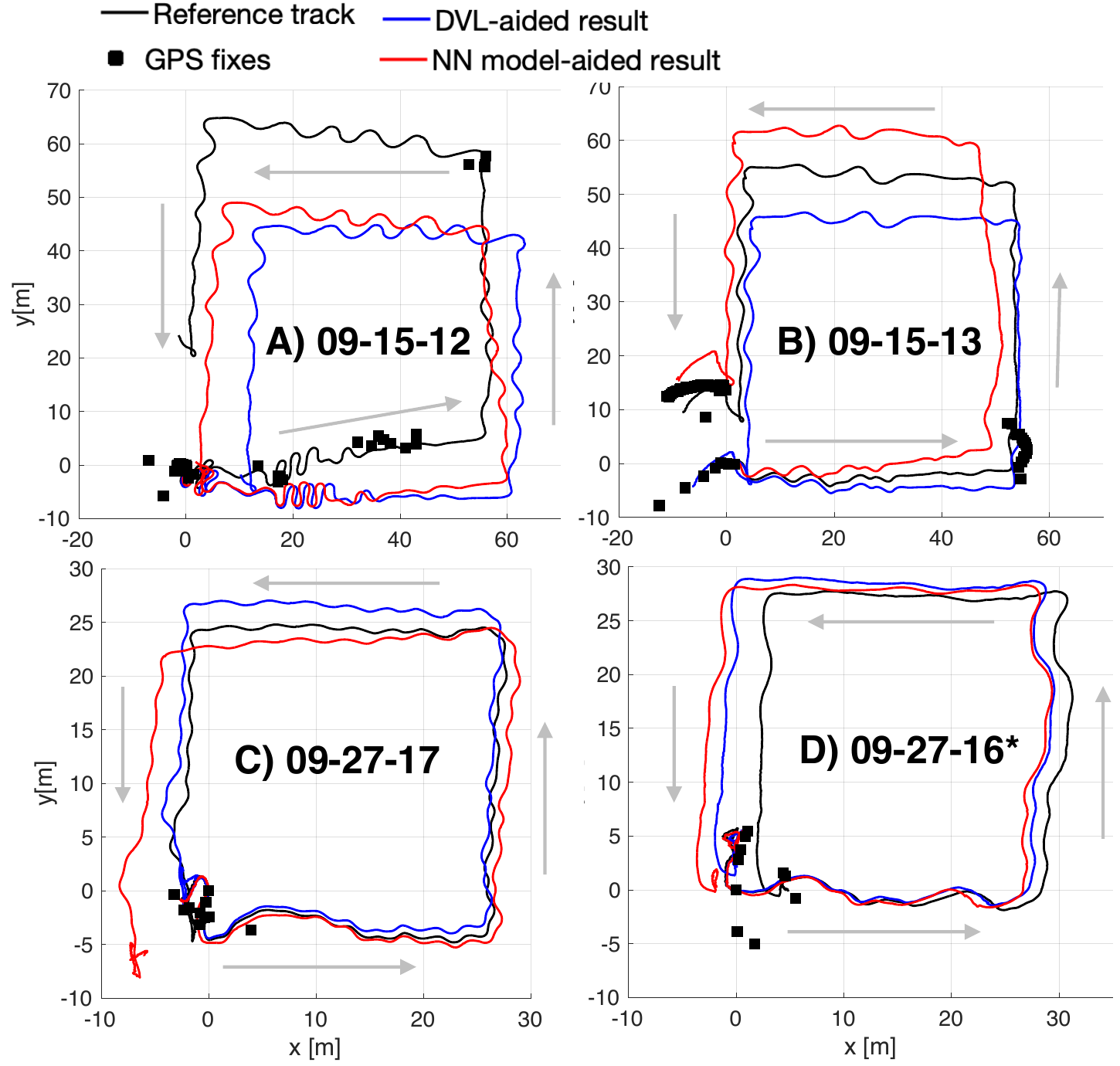}
    \caption{Comparison of the localization tracks from different methods. Black paths are the reference track generated by correcting the DVL-aided path (blue) with the GPS fixes (black squares). The red path is generated using the IMU data and the selected NN models. Some GPS fixes outliners are observed during the mission when the AUV is performing periodical surfacing. Note: the data from 27 September 2016 were used in model training and validation.}
    \label{fig:comparison}
\end{figure}

Fig.~\ref{fig:comparison} shows the comparison between the paths generated from the different localization methods. The black tracks are the reference track (current corrected using the GPS fixes shown in black squares). The blue paths are generated using the EKF in the robot localization package with DVL-measured velocities, IMU data, and pressure sensor data, which is a benchmark method that is widely used \cite{Zhao2022TowardsUS}. Finally, the red paths are generated using the same EKF, but the DVL measurements are replaced by the linear velocities predicted from the NN models. 

\begin{figure}[htbp]
    \centering
    \includegraphics[width=0.95\textwidth]{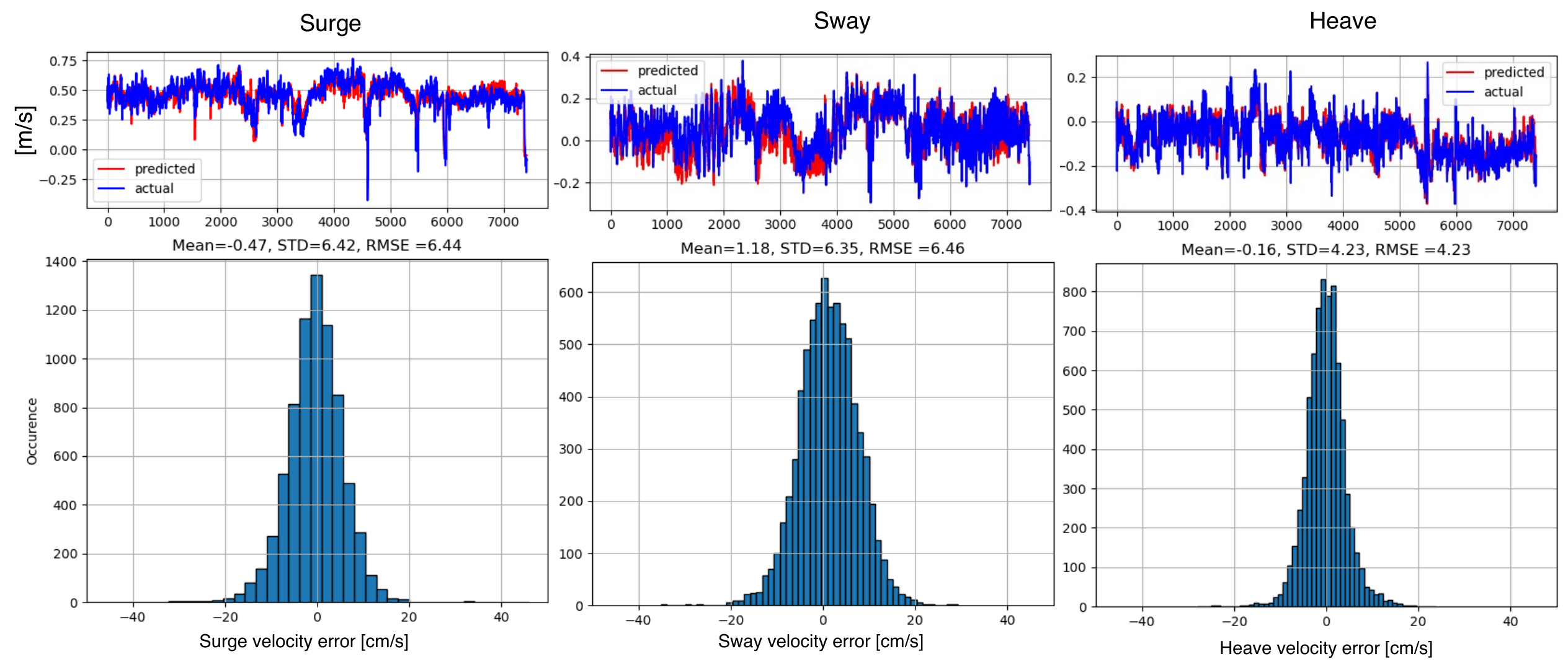}
    \caption{Comparison between the velocity predicted by the NN model and the velocity measured by the DVL. The error in surge and sway is increased compared to Fig.~\ref{fig:validation} because current effects were observed in the first two testing trials.}
    \label{fig:test_result}
\end{figure}

\begin{figure}[htbp]
    \centering
    \includegraphics[width=0.95\textwidth]{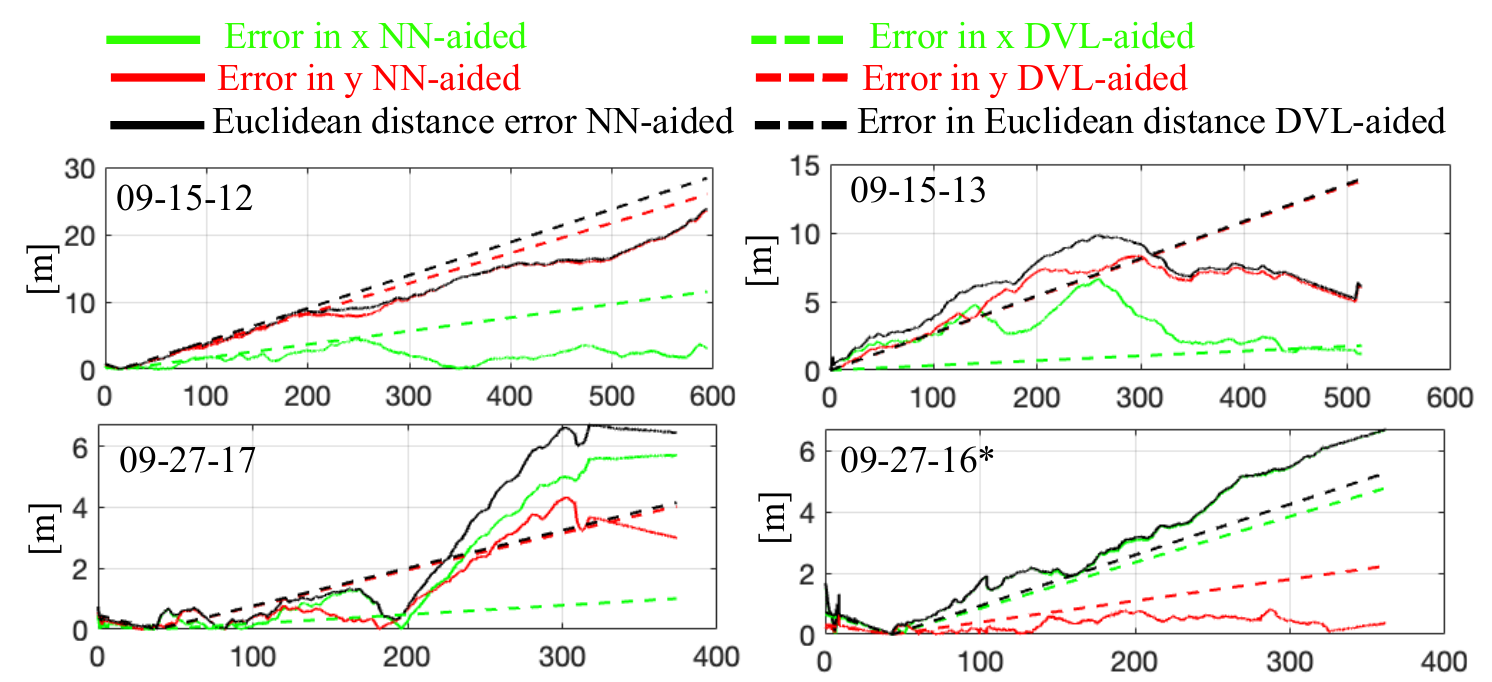}
    \caption{Comparison of the position error in $x$ (red) and $y$ (green), and Euclidean distance (black) for NN-aided (solid lines) and DVL-aided (dashed lines) methods. The two zoom-in plots are included for 27 September 2017 and 27 September 2016 since the range of errors is small. Note: the data from 27 September 2016 were used in model training and validation.}
    \label{fig:xy_error}
\end{figure}

Fig.~\ref{fig:test_result} presents the comparison of the velocities measured by the DVL (blue) and predicted by the NN models (red). We observe that the mean error in surge and sway is higher compared to the statistical values shown in Fig.~\ref{fig:validation}. This may be because the two test trials on 15 September were affected by a strong current (see Fig.~\ref{fig:comparison} and Table~\ref{tab:current_correction}). Furthermore, Fig.~\ref{fig:xy_error} presents the error in the x- and y-directions and the 2-D Euclidean distance between the navigation tracks (NN-aided and DVL-aided) and the reference track. We found that the NN-aided navigation tracks (solid lines) on 15 September both yielded smaller drifts compared to DVL-aided tracks (dash lines). In contrast, the DVL-aided navigation outperforms the NN-aided approach with the data from 27 September. However, the errors are significantly smaller on 27 September due to lower current correction compared to the data on 15 September. We believe the condition of the ocean current may have more impact on the DVL-aided localization compared to the NN-aided, which is trained using the data sets collected in different ocean conditions. 

We present the quantitative result in Table~\ref{tab:performance_table} with the root-mean-square (rms) distance error and the max distance error for each trial with/without DVL measurements. We also have normalized the values with respect to the total traveling distance, as we expect that the rms distance error and the maximum error will increase as a function of distance, but the ratio between them should be consistent. For estimating long-term navigation error, we have computed the drift per hour from each mission (about 15 min) and the maximum error per distance traveled. Both metrics are computed by dividing the maximum distance error over the mission time and traveling distance, and we found the average drift is about 100 m per hour (or 5\%--10\% drift for traveling distance) for both DVL-aided and NN-aided methods. The final distance error is not used in our analysis since it may be smaller than the maximum error when running box or repeated tracks, therefore, underestimating the navigation errors. Comparing the NN-aided and DVL-aided results, we found that the NN-aided trials have a lower drift rate, especially in the trials conducted on 15 September. However, as mentioned earlier, in the two trials on 27 September, the DVL-aided method outperforms the NN-aided method. Next, we compared the standard deviation from the two methods. The NN-aided method has a much smaller standard deviation in drift rates between different missions, meaning possibly a more consistent performance. As mentioned earlier, the water current impact is more significant on the trials conducted on 15 September, which may be the cause of the increased drift listed in Table~\ref{tab:performance_table} compared to the other three missions. We think the improved drift rate with the NN-aided method may be due to the inclusion of voltage and current as inputs for the NN models such that the NN model has some knowledge about the actual thrust generated from the thruster (a function of current and voltage supplied to the motor) for a better speed estimation over the ground, even under the impacts of ocean current. As a result, the NN-aided method outperforms DVL-aided when the AUV is experiencing a strong current. 

\begin{table}[htbp]
    \caption{Summary of the performance for the NN model-aided and the DVL-aided localization}
    \centering
    \begin{tabular}{c|c|c|c|c|c|c|c}
    \hline
    &Values &09-15-12&09-15-13&09-27-17&09-27-16*&Mean&STD\\
    \hline
    \multirow{5}{*}{NN-aided}&RMS distance error [m]&12.58&6.79&3.66&3.66&7.68&4.53\\
    \cline{2-8}
    &Max distance error [m]&23.84&9.89&6.70&6.68&13.48&9.12\\
    \cline{2-8}
    &RMSE/distance [\%]&4.41&2.86&2.61&2.63&3.29&0.97\\
    \cline{2-8}
    &Max error/distance [\%]&8.35&4.17&4.78&4.79&5.77&2.26\\
    \cline{2-8}
    &Max. drift per hour [m]&144.57&69.40&64.59&66.65&92.85&44.85\\
     \hline
     \multirow{5}{*}{DVL-aided}&RMS distance error [m]&16.23&8.06&2.28&2.85&8.86&7.01\\
    \cline{2-8}
    &Max distance error [m]&28.48&13.97&4.15&5.26&15.53&12.24\\
    \cline{2-8}
    &RMSE/distance [\%]&5.23&3.40&1.62&2.91&3.42&1.80\\
    \cline{2-8}
    &Max error/distance [\%]&9.98&5.89&2.96&3.77&6.28&3.53\\
    \cline{2-8}
    &Max drift per hour [m]&172.68&98.01&40.03&52.52&103.57&66.50\\
     \hline
    \end{tabular}

\smallskip
\footnotesize Note*: 80\% of the data from trial 27 September 2016 are used for training, while the data from other trials were only used for testing. Trial 27 September 2016 is excluded when computing the mean and standard deviation.
    \label{tab:performance_table}
\end{table}

Compared to other methods, our drift rate is comparable to the results presented in \cite{SupunAT2018ImplementationOA} and \cite{Arnold2018RobustMI}. However, as discussed in Section~\ref{sec:I}, Supun et al. \cite{SupunAT2018ImplementationOA} required an LBL and some vehicle-specific measurements and derivatives for speed prediction, and Arnold and Medagoda \cite{Arnold2018RobustMI} used a more expensive KVH IMU. In contrast, our approach requires a less expensive IMU and was tested on a relatively slow-moving AUV (at about 0.5 m/s with potentially a higher signal-to-noise ratio) without any acoustic localization systems for generating the reference data for model training.

\section{Conclusion and Future Works}\label{sec:VI}

This article presented a new method of using fully connected FNN to predict vehicle linear velocity components based on IMU data (acceleration, angular velocity, and orientation), pressure sensor (depth), thruster commands, system voltage, and current draws. The design, training, and selection of the FNN models were discussed. The most effective models were selected and fused with the IMU measurement for odometry calculation in playback mode using the EKF from robot localization \cite{MooreStouchKeneralizedEkf2014} based on a kinematics model without knowing the vehicle dynamics. 

We found that using the velocity components predicted from FNN models (up to two hidden layers and eight neurons in each layer) could produce a meaningful track. These lightweight NN models can be easily implemented on embedded processors commonly found onboard AUVs. We further compared the performance with the reference track (current corrected using GPS fixes) and the original localization method using DVL measurements, IMU, and pressure sensor. We found that the new method (using velocity predicted from the FNN) has a slightly lower and more consistent drift rate compared to the path generated using the DVL velocity measurements, with an overall drift rate per hour slightly under 100 m. The findings suggest that, with sufficient training data, the FNN models could help provide relatively good vehicle velocity estimates when DVL is not available either because of budget constraints or because the vehicle is operated away from the bottom. However, our method still needs a DVL to collect the training data, similar to existing model-based localization methods \cite{SupunAT2018ImplementationOA}, \cite{Arnold2018RobustMI}. Once the models are trained, the DVL can be removed during the real deployment, but the method is still subjected to error growth without bounds (even a slower rate compared to the DVL-aided method) since no external aids, such as a USBL system, are available. 

In the future, we plan to further include water track velocity measurements as the inputs of the NN models to better capture the impacts from currents. In \cite{Arnold2018RobustMI}, improvements have been found when including ADCP measurements. However, most DVLs do not offer water track measurement or are offered with extra costs, and the new NN model will require the DVL or ADCP staying on the AUV, which will increase the overall vehicle cost. We also have planned to conduct more field trials with longer mission periods to further investigate the robustness of the proposed method and the impacts of the ocean current on the NN-aided method. On the one hand, longer mission time in different path patterns will reveal underfitting and overfitting scenarios in our NN models. On the other hand, we found that the improved robustness of the NN-aided localization under strong currents needs further investigation. While our initial conclusion is that due to the inclusion of the system voltage and current draw in the NN models that could provide some knowledge about the actual thrust from the thrusters, more data will be needed to prove the hypothesis. One current limitation of the approach is that the NN models, which is approximating the vehicle dynamics, need to be retrained if a different vehicle is used. We plan to investigate the potential of creating an online adaptive mechanism for model adaption on different vehicles. The mechanism (an NN or a parameterized nonlinear function) will be modeled only using a short period of motion data from the target vehicle and output scaling factors to remap the velocity outputs from pretrained NN models. The adaptive method will reduce the overall time in training the NN models when a new vehicle is used or there are significant physical changes on a existing vehicle.

\section*{Acknowledgment}
The authors would also like to thank Brian Caccioppoli for operating the small boat for the field data collection.

\begin{IEEEbiographynophoto}{Mingxi Zhou}
(Member, IEEE) received the Ph.D. degree in ocean and naval architectural from the Memorial University of Newfoundland, St. John's, NL, Canada, in 2017. He is currently an Associate Professor with the Graduate School of Oceanography, University of Rhode Island, Kingston, RI, USA. His current research interests include underwater navigation, AUV system development, and autonomy.
\end{IEEEbiographynophoto}

\begin{IEEEbiographynophoto}{Emir Cem Gezer}
received the B.Sc. degree in computer engineering from Ankara University, Ankara, T{\"u}rkiye, in 2019, and the M.Sc. degree in ocean engineering from the University of Rhode Island, Kingston, RI, USA, in 2022. He is currently working toward the Ph.D. degree in risk-aware and safeguarding control for autonomous ships with the Norwegian University of Science and Technology, Trondheim, Norway.
\end{IEEEbiographynophoto}

\begin{IEEEbiographynophoto}{Lin Zhao}
(Member, IEEE) received the Ph.D. degree in ocean engineering from the University of Rhode Island, Kingston, RI, USA, in 2024. Since 2024, he has been a Postdoctoral Fellow in electrical, computer, and biomedical engineering, working on multisensor fusion SLAM for single-and multirobot marine systems. He is currently a Senior Perception Engineer with Anyware Robotics, Fremont, CA, USA.
\end{IEEEbiographynophoto}

\begin{IEEEbiographynophoto}{William McConnell}
received the B.Sc. degree in mechanical engineering from the University of Rhode Island, Kingston, RI, USA, in 2024. He was an undergraduate Researcher at the time of his contribution to this article, and is currently a Mechanical Engineer and Product Lead with Vatn Systems, Bristol, RI, USA.
\end{IEEEbiographynophoto}

\begin{IEEEbiographynophoto}{Chengzhi Yuan}
(Member, IEEE) received the Ph.D. degree in mechanical engineering from North Carolina State University, Raleigh, NC, USA, in 2016. He is currently an Associate Professor with the Mechanical, Industrial and Systems Engineering Department, University of Rhode Island, Kingston, RI, USA. His current research interests include adaptive learning control theory and its applications to robotic systems.
\end{IEEEbiographynophoto}

\end{document}